\documentclass[sigconf]{acmart}

\copyrightyear{2021} 
\acmYear{2021} 
\setcopyright{acmcopyright}
\acmConference[SIGMOD '21]{Proceedings of the 2021 International Conference on Management of Data}{June 20--25, 2021}{Virtual Event, China}
\acmBooktitle{Proceedings of the 2021 International Conference on Management of Data (SIGMOD '21), June 20--25, 2021, Virtual Event, China}
\acmPrice{15.00}
\acmDOI{10.1145/3448016.3457554}
\acmISBN{978-1-4503-8343-1/21/06}

\usepackage{xcolor}

\AtBeginDocument{%
  \providecommand\BibTeX{{%
    \normalfont B\kern-0.5em{\scshape i\kern-0.25em b}\kern-0.8em\TeX}}}

\newtheorem{definition}{Definition}

\newcommand{\eat}[1]{}

\begin{document}
\fancyhead{}
\title{Learning-Aided Heuristics Design for Storage System}

\author{Yingtian Tang$^{1,3}$, Han Lu$^{1,2}$, Xijun Li$^{1,4}$, Lei Chen$^{1}$, Mingxuan Yuan$^{1}$ and Jia Zeng$^{1}$}

\authornote{Han Lu and Yingtian Tang contributed equally to this research. This work was done when Han and Yingtian were interns in Huawei Noah's Ark Lab. Xijun Li is the corresponding author.}
\affiliation{%
	\institution{1 Huawei Noah's Ark Lab}
	\institution{2 Shanghai Jiao Tong University}
	\institution{3 University of Pennsylvania}
	\institution{4 University of Science and Technology of China}
}


\begin{abstract}
    Computer systems such as storage systems normally require transparent white-box algorithms that are interpretable for human experts. In this work, we propose a learning-aided heuristic design method,  which automatically generates human-readable strategies from Deep Reinforcement Learning (DRL) agents. This method benefits from the power of deep learning but avoids the shortcoming of its black-box property. Besides the white-box advantage, experiments in our storage production’s resource allocation scenario also show that this solution outperforms the system’s default settings and the elaborately handcrafted strategy by human experts. 
\end{abstract}




\maketitle
\section{Introduction}

\textcolor{black}{The tuning process of computer systems is laborious and expensive; but of great significance in terms of performance. Effective and explainable heuristics have been widely adopted for tuning in existing computer systems. For example, conventional heuristics such as FIFO (First In First Out) and LRU (Least Recently Used) are widely used in cache scenario. Nevertheless, these strategies are customized and require sophisticated handcraft design. Recently, a promising field of using machine learning to optimize computer systems is drawing increasing attentions~\cite{maas2020taxonomy}. For example, Mao \textit{et. al.}~\cite{mao2016resource} apply deep reinforcement learning (DRL) on resource management problems. Kraska \textit{et. al.}~\cite{kraska2018case} attempt to replace traditional index in computer system (such as B-Tree and BitMap) by learning the mapping relation. However, the black-box property hinders the deployment of them in real-world systems, where thorough sanity checks and complete traceability are required.  
There have been attempts to interpret the deep learning models, for instance, Koul \textit{et. al.}~\cite{koul2018learning} propose a method to extract semantic strategies from a trained recurrent neural network and uses that to play Atari games. However, how these methods could be used for practical computer system tuning still remains underexplored.}



In this paper, we consider a real-time resource allocation problem in the storage system Dorado V6~\cite{dorado_v6}. The computation resources (\textit{i.e.}, CPU cores) are expected to be appropriately allocated into different levels when workloads dynamically change, so as to finish a given amount of IO requests with the minimum processing time. 
Considering limited computation resource for tuning and the system performance safety issue, a lightweight white-box approach is required. To take advantage of the power of deep reinforcement learning, we propose a learning-aided heuristics design method which can extract an explainable finite state control strategy.


Our contributions are summarized as follows:
\begin{itemize}
    \item We present an integrated pipeline of learning heuristics from DRL policies, including heuristics extraction, generalization capability enhancement, and interpretation, which aims to facilitate domain experts to devise more sophisticated heuristics for computer systems.
    \item We apply the above methods on a real-time resource allocation scenario in our storage product. Experimental results show that both the DRL model and corresponding extracted heuristics outperform 
    the default production setting and the elaborately handcrafted strategy by human experts. 
\end{itemize}

\vspace{-1.7ex}
\section{Problem description}
\label{sec: problem description}

\begin{figure}[t]
	\centering
	\includegraphics[width=0.9\linewidth]{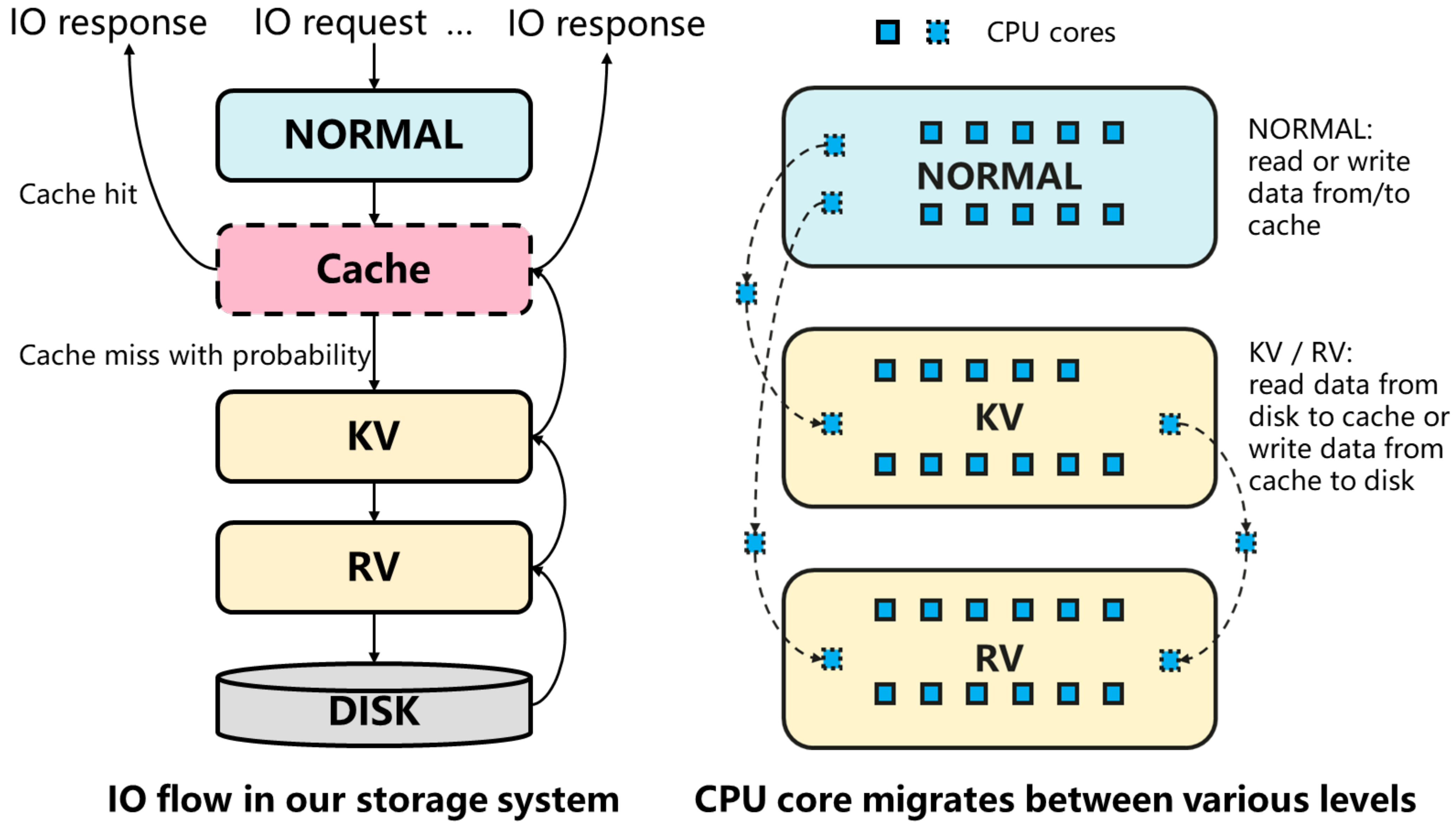}
	\caption{CPU core allocation problem in our storage system} 
	\label{fig:problem_description}
\end{figure}

As shown in Figure~\ref{fig:problem_description}, Dorado V6 storage system has a multi-level computation resource architecture. There exist three levels where CPU cores can reside, \textit{i.e.}, NORMAL, KV and RV. \textcolor{black}{Specifically, KV level stands for Key-Value storage level, in which CPU cores are utilized to calculate the key-value mapping relation. RV level stands for Resource Volume level, in which CPU cores are used in virtualization management of disk resources.}
The CPU cores in different levels perform different duties. The cores in NORMAL level load data from a shared cache. The cache miss occurs when the NORMAL level cache does not contain the requested data. In this case, the CPU cores in KV and RV levels fetch data from the disk and load it into the cache of NORMAL level. The cores in NORMAL level can then read the loaded data from cache to meet the IO request. \textcolor{black}{There are various kinds of IO requests, different in the size and type of read/write. Note that read IO requests might be finished only with CPU cores of NORMAL level, while for write IO requests, cores of all three levels must be used.} For any time interval, workloads that comprise different kinds of IO requests are concurrently sent to the storage system.
Our goal is to migrate CPU cores among the three levels according to the dynamic workload distributions, to finish the workload with the minimum number of time intervals. The formal definitions are given as follows.

\vspace{-2ex}
\begin{definition}[Workload]
Workload $w(t)$ within one time interval $t$ can be described with two 14-dimension vectors and a scalar:
\begin{equation}
    S_w(t) = [S_i], \quad i=1,2,..., 14
\end{equation}
\begin{equation}
    I_w(t) = [I_i], \quad i=1,2,..., 14
\end{equation}
\begin{equation}
    Q_w(t)
\end{equation}
where $S_i$ and $I_i$ together describe the $i$th type of IO request. $S_i$ denotes the IO size and type (read/write). $I_i$ denotes the ratio of $S_i$ in $w$. Therefore, $\sum_{i=1}^{14} I_i=1$. $Q_w(t)$ represents the total number of IO requests in workload $w(t)$ sent to storage system within time interval $t$.

\end{definition}

\vspace{-3ex}
\begin{definition}[Maximum Processing Capability]
 Each CPU core has its own maximum processing capability per time interval, denoted as $m$, which is the maximum sum of IO request sizes (not the number of IO requests) that could be processed. Suppose that the total number of cores in the storage system is $N$, the ideal maximum processing capability within one time interval is $N\times m$.
\end{definition}
\vspace{-2ex}
\begin{definition}[Cache Miss Rate]
 CPU cores of NORMAL level may fail to read data from their own cache, which is referred to as cache miss. The probability of cache miss is denoted as $C$.
\end{definition}
\vspace{-1ex}
Besides the above definitions, the storage system also has the following properties: 1) IO requests of workload $w$ are assigned to cores in a polling manner; 2) Each IO request must be processed and cannot be discarded. If an IO request sent in a time interval $t$ cannot be processed (\textit{e.g.,} exceeds the total processing capability of all cores within time interval $t$), then it will be postponed to the following time intervals; 3) Each core is able to migrate between NORMAL, KV, and RV levels, as shown in Figure~\ref{fig:problem_description}. A core must finish all the IO requests assigned to it before migration. 
A certain percentage of performance loss in the next time interval would be caused by the migration of a core.


\noindent\textbf{Objective:} For a sequence of workloads $w(t|t=1,2,..T)$, the makespan $K$ ($K \geq T$) is the number of time intervals to finish all IO requests.
Our objective is to design dynamic CPU migration policies among different levels to minimize the makespan $K$.

\section{Heuristics Learned from DRL Policy}
\begin{figure}[t]
	\centering
	\includegraphics[width=0.9\linewidth]{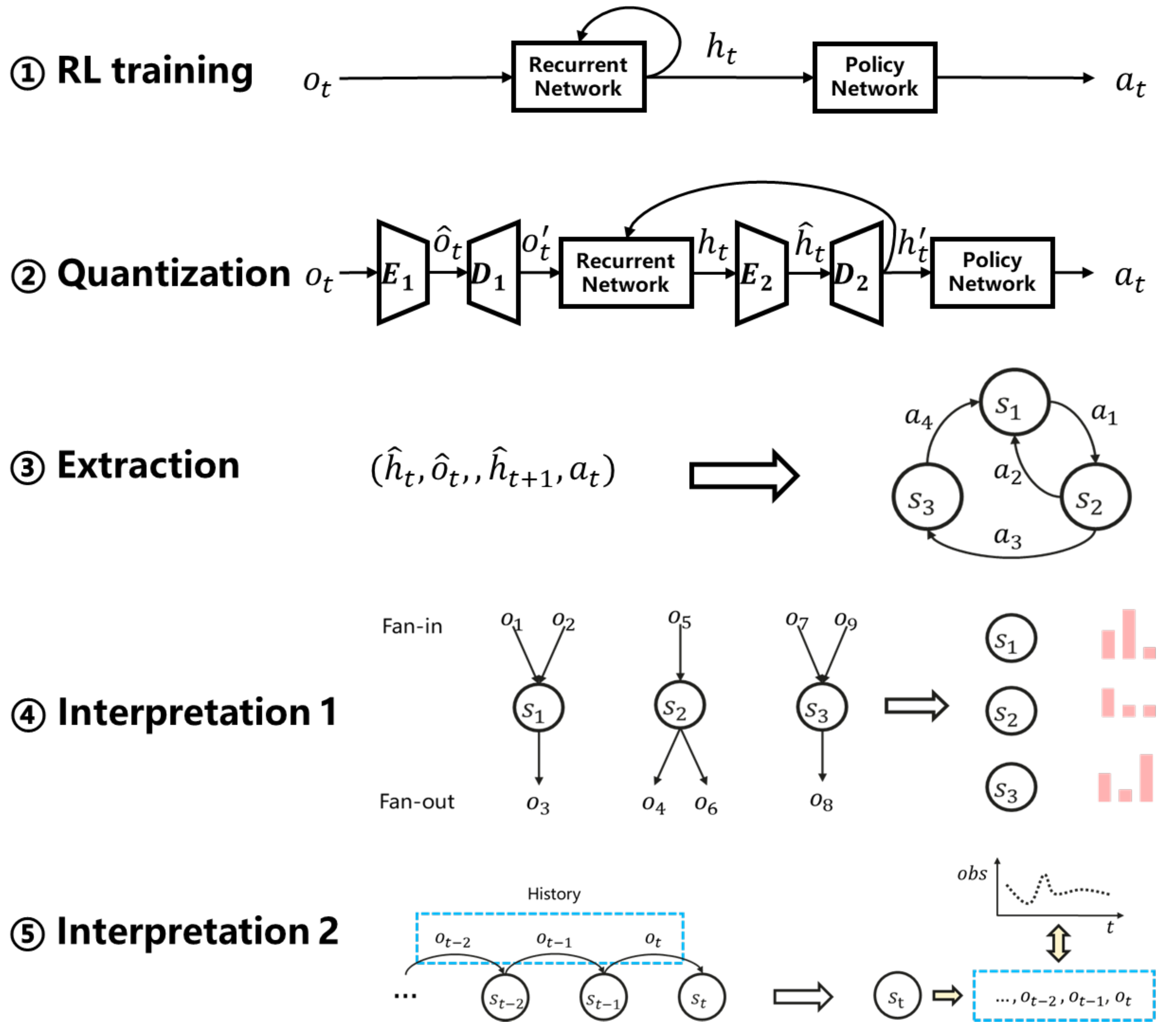}
	\caption{The detailed process of our proposed method to automatically extract heuristics using RNN-based reinforcement learning, where a QBN~\cite{koul2018learning} technique is adopted in the first three steps.}
	\label{fig:solution}
\end{figure}

In this section, we present an integrate pipeline that learns heuristics from DRL policies. The overall process is shown in Figure~\ref{fig:solution}. We first construct a recurrent DRL model that consists of a value network and a policy network, and train it in the environment. Then we insert two quantization auto-encoders and retrain the model. The auto-encoders are for the observed data and the hidden states, respectively. Next, we extract a Finite State Machine (FSM) from the embedding of the auto-encoders. Finally, we summarize semantics of the FSM states via matching corresponding observations and analyzing them statistically.

\vspace{-1.5ex}

\subsection{RNN-based Reinforcement Learning}
We model the problem as a Markov Decision Process (MDP), then utilize a Recurrent Neural Network (RNN) based DRL model to solve the MDP. It is believed that a latent context relation exists in the transitions of workloads. The RNN in our DRL model is devised to cope with this latent relationship.

\noindent\textbf{Observation:} $o_t$ represents the observation at the time interval $t$. $o_t=[c_N(t),c_K(t),c_R(t),u_N(t),u_K(t),u_R(t),w(t),Q_w(t)]$, where $c_N(t)$, $c_K(t)$ and $c_R(t)$ respectively denote the number of cores in NORMAL, KV, RV levels and $u_N(t)$, $u_K(t)$ and $u_R(t)$ respectively represent the average CPU utilization rate of the three levels. 
The observation space is denoted as $\mathcal{O}=\{o_t\}$.

\noindent\textbf{Hidden State:} $h_t$ denotes the hidden state at time interval $t$, which is updated on each transition and affects the following action selection. In particular, $h_t=\phi (h_{t-1}, o_t)$, where $\phi$ is the transition function maintained in the recurrent network. The hidden state space is denoted by $\mathcal{H}$. Hence the transition function $\phi$ is a mapping: $\mathcal{H} \times \mathcal{O} \mapsto \mathcal{H}$.

\noindent\textbf{Action:} For each hidden state $h_t$, an action $a_t$ is chosen from the policy network $\pi$, \textit{i.e.,} $a_t=\pi(h_t)$. The action space is denoted as $\mathcal{A}=\{a_i|i=1,...,7\}$, where there are seven distinct actions in total. Note that action $a_1$ represents no CPU core migration between different levels. The rest of actions respectively denote migrating one CPU core from one level to another level (\textit{e.g.}, migrating one core from NORMAL to KV).

\noindent\textbf{Reward:} 
The reward is measured as $1/K$, which is the inverse of the makespan.

\vspace{-2.5ex}

\subsection{Extract Heuristics from Learned Policy}

\subsubsection{Finite State Machine Extraction}
 As Step 2 and 3 illustrate in Figure~\ref{fig:solution}, once the DRL model converges in training, we extract a FSM from it with Quantized Bottleneck Network (QBN) technique~\cite{koul2018learning}. The QBNs are auto-encoders that reconstruct continuous embedded observations $o_t$ and hidden states $h_t$ as $o_{t}^{'}$ and $h_{t}^{'}$. The entries of their latent embeddings $\widehat{o}_t$ and $\widehat{h}_t$ (whose dimension is denoted as $L$) are restricted to be $k$-bit quantized. There are $k^L$ distinct embeddings that span the discrete embedding space $\mathcal{\widehat{O}}$ and $\mathcal{\widehat{H}}$. A dataset of $<h_{t}, h_{t+1}, o_t, a_t>$ can be collected via running the trained DRL model. The QBNs are then trained over the collected dataset using supervised learning to minimize the reconstruction error. In this way, a discrete dataset of $<\widehat{h}_{t}, \widehat{h}_{t+1}, \widehat{o}_{t}, a_t>$ is obtained, which produces a transition table, \textit{i.e.,} the extracted FSM.


\vspace{-1ex}
\subsubsection{Generalization Capability Enhancement} 
Unlike the classical video game scenarios~\cite{koul2018learning} where the DRL and extracted FSM could see all possible observations, in our scenario we cannot observe all possible kinds of workloads. We thus propose two methods to enhance the generalization capability of the extracted FSM. 

The first one is \textit{curriculum learning}. In practice, we cannot operate the storage system once it has been sold to our customers. Thus we are not able to obtain large amounts of real workload traces from the users unless we are granted with the permissions. As the result, only a few real workload traces are available to us. However, we can collect summarized characteristics of real workload traces, such as periods, trends and dominant IO types, via customer investigation (a common business mode). With these characteristics, we construct several standard workload traces using Vdbench~\cite{vdbench}. 
We regard standard workload traces as \textit{easy tasks}. A policy $\pi_{\theta}$ is first trained on multiple easy tasks until it converges. The real workload traces are regarded as \textit{hard tasks}, the number of which are very few compared to the number of easy tasks. With knowledge learned in easy tasks, we continue to train policy $\pi_{\theta}$ on a few hard tasks to get the final policy. We experimentally validate that the proposed method improves the generalization capability in Section~\ref{sec: experiment}. 

The second one is to classify an unseen observation as its closest known observation. The intuition behind is that the state space has a certain continuity and similar observations could trigger similar actions.
Specifically, we define the ``closeness'' of two observations as the similarity between their observation vectors. The similarity measures such as \textit{Euclidean} distance and cosine similarity can be applied. The unseen observation can therefore trigger a transition in the extracted FSM.

\vspace{-2ex}

\subsection{Interpretation of Extracted States}
\label{subsection: interpretation}



\textcolor{black}{A FSM for real workload could be extracted from the trained DRL using the method mentioned above. We interpret the extracted states in two ways, so that the strategies of DRL could be unfolded for inspiring further heuristics design. \\
\indent We firstly examine the transitions into and from each state. Every state in the extracted FSM is associated with many observations which are divided into two classes, Fan-in and Fan-out, as shown in Figure~\ref{fig:solution} (observations that correspond to transitions between the same state should be ruled out). Moreover, each state corresponds to one unique action. It is the action emitted by the state that causes the variation between Fan-in and Fan-out observations. Here the original continuous observations are used instead of the quantized counterparts obtained from auto-encoders. For each state, we compare the average Fan-in and Fan-out observations, and infer how the state reacts to environment and the intensity of that reaction.\\
\indent Secondly, we examine the history of observations before the transition into a state. For each presence of a specific state, we collect a time window of observations happened before it. We then take the average of these time windows, which represents the general history information of that state. The states are extracted from the RNN-based DRL, so that the history that we obtained could be useful for explaining what causes the transition into a state and what information is encoded by the state. }

\vspace{-1.5ex}
\section{Experiment} \label{sec: experiment}
\subsection{Settings}

$12$ classes of standard workload traces are synthesized using the Vdbench tool, each of which is associated with one typical business model of the users, such as database, heavy computing, \textit{etc}. Recall that we only have very few real workload traces from the users. We simulate real workload traces by sampling snippets from the aforementioned standard workloads. In this way, we generate $50$ workload traces. To sample more efficiently for RL, we write a simulator to simulate the CPU core migration in Dorado V6 storage system. In addition to the characteristics of storage system described in Section~\ref{sec: problem description}, we also consider the idle rate of CPU cores which follows a \textit{Poisson} distribution in the simulator.

\vspace{-2ex}
\subsection{Training procedures}
We use a Gated Recurrent Unit (GRU) with $128$ hidden nodes to incorporate the recurrent architecture. We forward its hidden state to two linear layers, with output sizes of $7$ and $1$ respectively, to produce the logits corresponding to all possible actions and the predicted state value. The loss design follows the Advantage Actor-Critic method (A2C)~\cite{mnih2016asynchronous}. We use Adam~\cite{kingma2014adam} optimizer with an initial learning rate $0.0003$ and clip the norm of gradients to be under $2$. The RL learning follows the \textit{Epsilon} greedy exploration with $0.1$ as the probability of random action selection. We adopt the method in~\cite{koul2018learning} to extract a finite state machine from the trained DRL model. For the parameters of QBNs, we set $k=3$ and $L=64$.
\vspace{-2ex}
\subsection{Experimental results}

\subsubsection{Convergence}

\begin{figure}[t]
	\centering
	\includegraphics[width=0.83\linewidth]{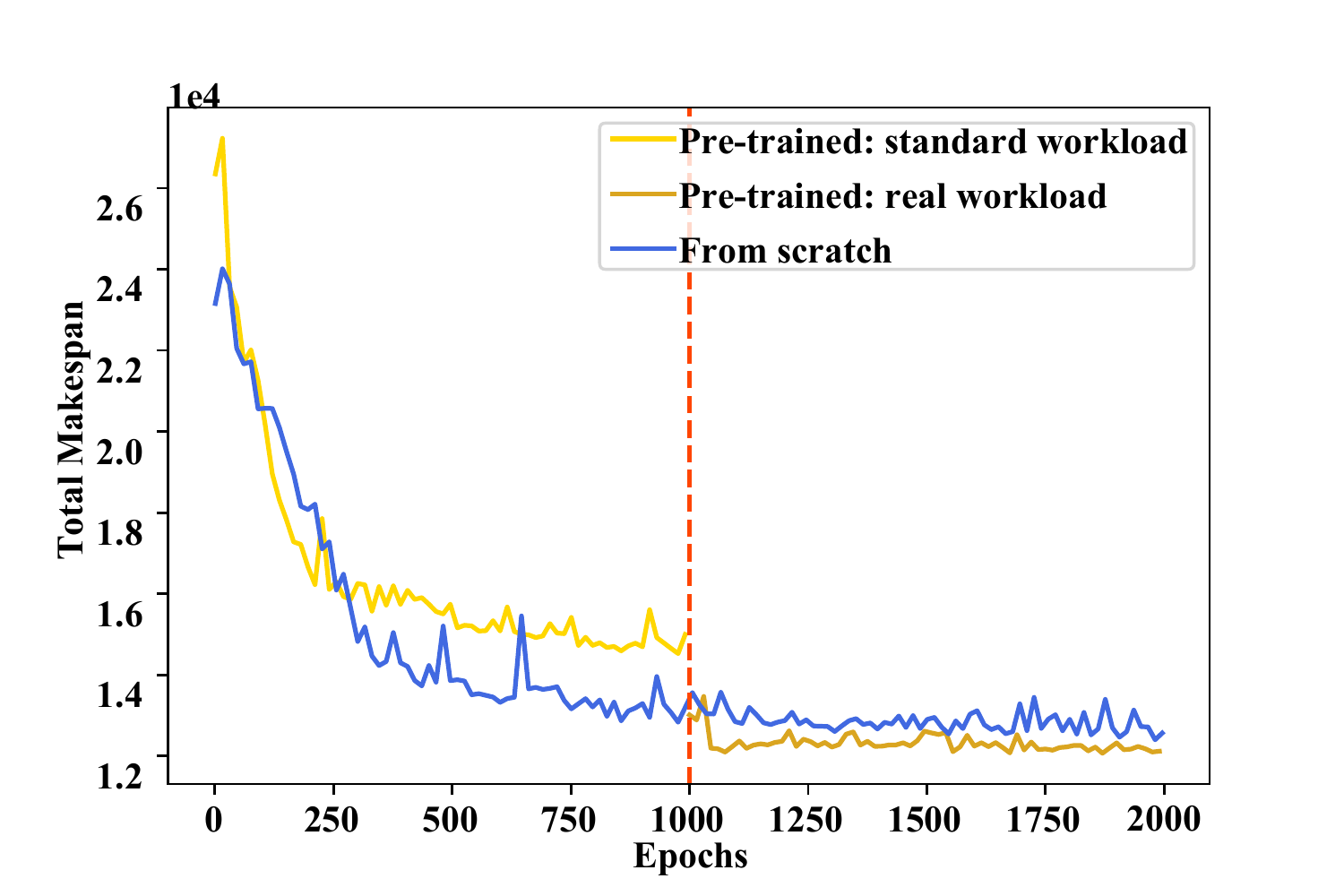}
	\caption{Convergence comparison}
	\label{fig: convergence}
\end{figure}

The proposed curriculum learning for storage system is validated here. One RL agent is trained for $2000$ epochs in total based on our curriculum learning strategy ($1000$ epochs for standard workload traces and $1000$ epochs for real workload traces). We train another agent on real workload traces for $2000$ epochs for comparison. The result of convergence comparison is shown in Figure~\ref{fig: convergence}. The blue curve represents the convergence process of training only on real workloads whilst the yellow and brown curves together show the convergence process of curriculum learning. The horizontal and vertical axes denote the epoch number and total makespan respectively. It can be seen that the RL agent with curriculum learning converges faster and better than the one learned from scratch. Besides, it is worth noting that the computation power consumption of training on standard workload traces is relatively lower than that on real workload traces. Thus, with the introduction of curriculum learning, we can obtain a better RL policy using lower computation power and less number of real workloads, which is vital to algorithm deployment in the production environment.

\vspace{-1.5ex}
\subsubsection{Performance} 
\begin{figure}[t]
	\centering
	\includegraphics[width=0.85\linewidth]{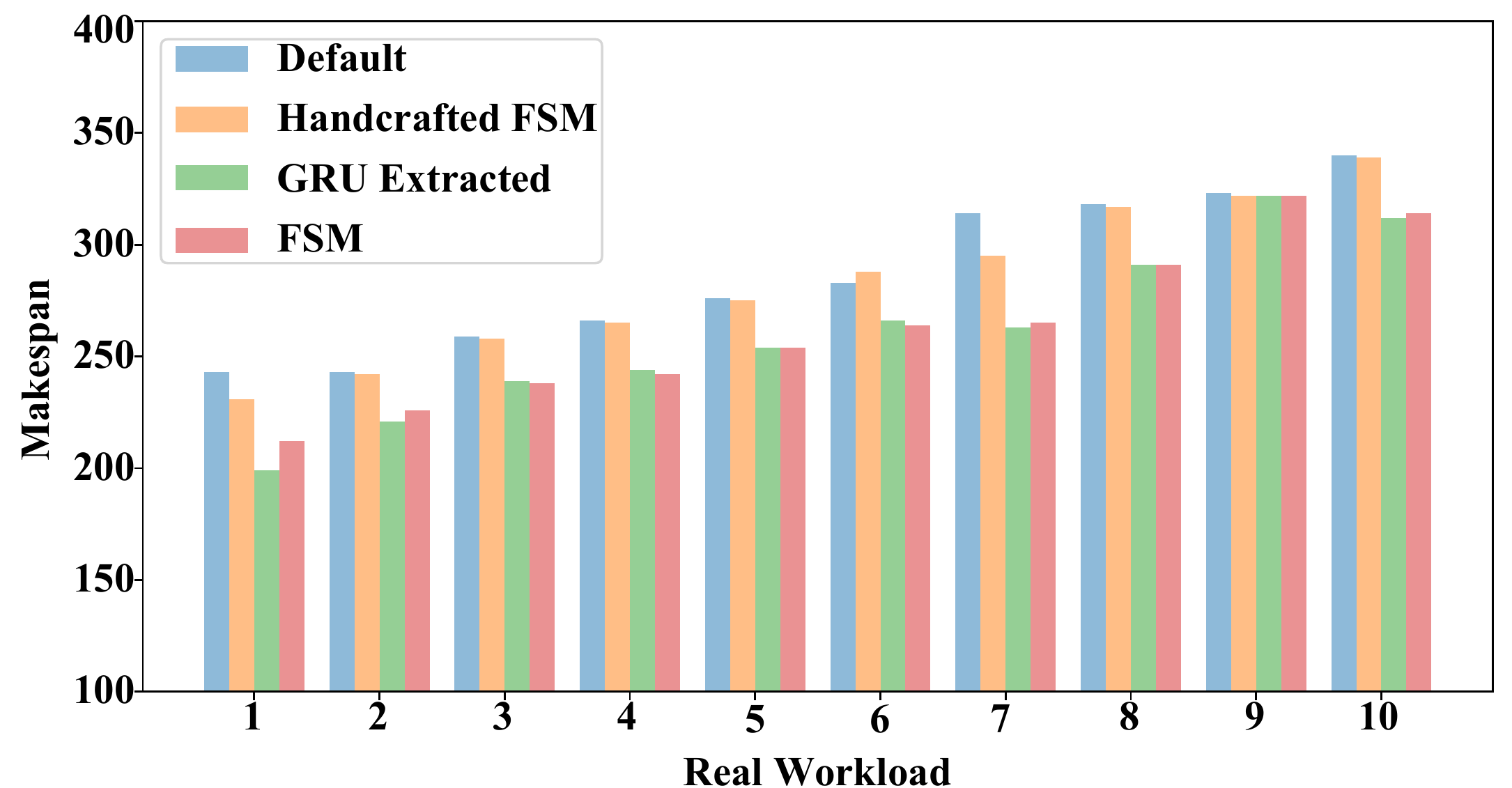}
	\caption{Performance comparison}
	\label{fig:accuracy comparison}
\end{figure}

Prior to interpret the extracted FSM, we are supposed to ensure that its behavior and performance are aligned with the original DRL model. We compare the performance between the original DRL model (\textit{i.e.,} GRU-based DRL) trained with curriculum learning, extracted FSM, a handcrafted FSM, and the default setting. Roughly speaking, the principle of handcrafted FSM is migrating CPU cores from the level with the lowest CPU utilization rate to the one with the highest CPU utilization rate. It is tested in User Acceptance Testing environment and show $20\%$ reduction of makespan. The default setting refers to no CPU migration during testing. The comparison result over ten real workloads is illustrated in Figure~\ref{fig:accuracy comparison}. It shows that all algorithms get lower average makespan than the default setting. Besides, both the original DRL model and extracted FSM perform better than handcrafted FSM ($11.5\%$ reduction of makespan on average). The extracted FSM performs a little bit worse than the original DRL model ($0.88\%$ increase of makespan on average) since there must be a loss of information after the quantizaion of DRL model. 


\vspace{-1.5ex}
\subsection{FSM interpretation}
\begin{figure}[t]
	\centering
	\includegraphics[width=0.85\linewidth]{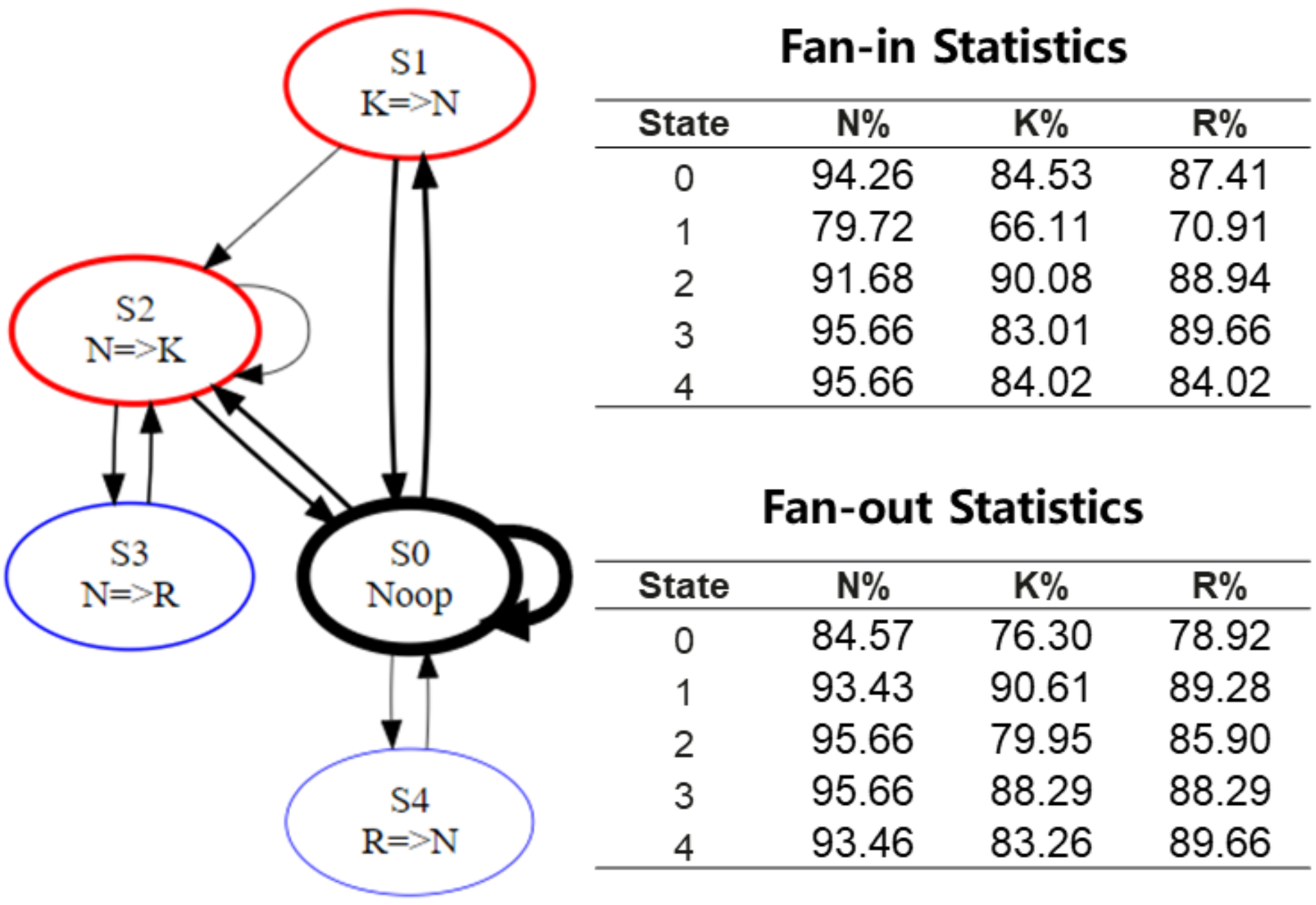}
	\caption{Visualization of extracted FSM}
	\label{fig:fsm_visual}
\end{figure}

\begin{figure}[t]
	\centering
	\includegraphics[width=0.85\linewidth]{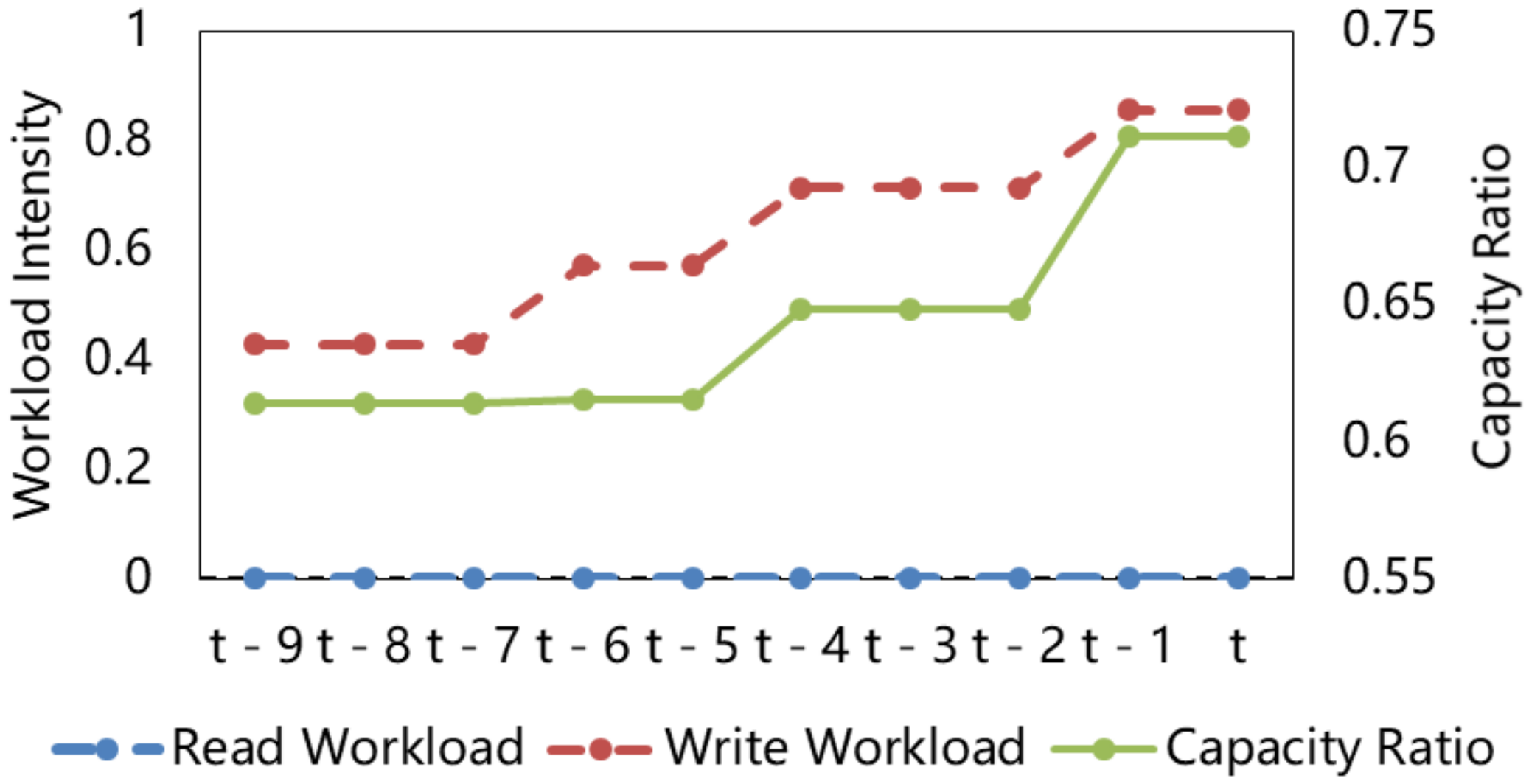}
	\caption{History information of the S2}
	\label{fig:history}
\end{figure}

\textcolor{black}{We visualize an extracted FSM over a real workload in Figure~\ref{fig:fsm_visual}. There are in total five states (\textit{i.e.,} the circles) in the FSM, where each state is associated with an action. For examples, ``Noop'' stands for no operation, \textit{i.e.,} without CPU core migration. ``N=>R'' refers to migrating one core from NORMAL level to RV level. The thickness of circle denotes how many transitions are associated with the state when applying the extracted FSM to the real workload.\\
\indent Using the method described in Section~\ref{subsection: interpretation}, we analyze the Fan-in and Fan-out statistics and the history information. The Fan-in and Fan-out statistics in Figure~\ref{fig:fsm_visual} illustrate the basic semantic meaning of each state. S0 (``Noop'') is the most frequent state, since the FSM adapts to the workload and then keeps the stabilized configuration in the long term. The difference between its Fan-in and Fan-out CPU utilization indicates the general fluctuation of workload intensity. For S1 and S4, they tend to move cores from the level with low utilization to the levels with high utilization. This is a basic strategy that simply gives the level with high demand more computation capacity, which is also the strategy used by our handcrafted FSM. \\
\indent S2 and S3 do not follow this basic strategy. The history information of S2 in Figure~\ref{fig:history} explains this phenomenon. The figure shows information of the last 10 average observations before the transition into S2. Recall that read requests only demand loading the data, whereas write requests additionally require writing data back to disk. Here we see that the intensity of write workload keeps rising while the intensity of read workload stays at 0. Moreover, the capacity ratio (the ratio of computation capacity of NORMAL to that of KV and RV) goes up. Obviously, the FSM tried to firstly load all relevant data by increasing the capacity of NORMAL. At this moment, it readjusts to give KV and RV more capacity so that the write-back phase of write requests could be satisfied quickly. S3 also has similar history. We do not show it due to the limited space.\\
}

\vspace{-2ex}
\section{Conclusion}
In this paper, an integrated pipeline of learning heuristics from DRL policies is presented. We apply the proposed methods to a practical resource allocation problem in our storage product. Experimental results demonstrate that both the DRL model and extracted FSM outperform handcrafted FSM on various workloads. Visual and statistical analyses of the extracted FSM are given to provide insights into the DRL model for domain experts.

\bibliographystyle{ACM-Reference-Format}
\bibliography{sample-base}

\end{document}